\documentclass{article}
\usepackage{makecell}

\usepackage{graphicx}
\usepackage{wrapfig}

\usepackage[final]{neurips_2025}

\usepackage[utf8]{inputenc}
\usepackage[T1]{fontenc}
\usepackage{url}
\usepackage{booktabs}       
\usepackage{amsfonts}
\usepackage{nicefrac}
\usepackage{microtype}
\usepackage{xcolor}
\usepackage{tabularx}
\usepackage{ragged2e}
\usepackage{array}
\usepackage{listings}
\usepackage{multirow}

\title{Watch Wider and Think Deeper: Collaborative Cross-modal Chain-of-Thought for Complex Visual Reasoning}

\author{%
  Wenting Lu\textsuperscript{*} \\
  Fujian Normal University \\
  \texttt{qsx20231369@fjnu.edu.cn} \\
  \and
  Didi Zhu\textsuperscript{*} \\
  Zhejiang University \\
  \texttt{didi\_zhu@zju.edu.cn} \\
  \and
  Tao Shen \\
  Zhejiang University \\
  \texttt{tao.shen@zju.edu.cn} \\
  \and
  Donglin Zhu \\
  Zhejiang Normal University \\
  \texttt{donglin@zjnu.edu.cn} \\
  \and
  Ayong Ye\textsuperscript{\dag} \\
  Fujian Normal University \\
  \texttt{yay@fjnu.edu.cn} \\
  \and
  Chao Wu\textsuperscript{\dag} \\
  Zhejiang University \\
  \texttt{chao.wu@zju.edu.cn} \\
}

\begin{document}

\maketitle
\renewcommand{\thefootnote}{\fnsymbol{footnote}}
\footnotetext[1]{These authors contributed equally to this work}
\footnotetext[2]{Corresponding authors}

\begin{abstract}
Multi-modal reasoning requires the seamless integration of visual and linguistic cues, yet existing Chain-of-Thought methods suffer from two critical limitations in cross-modal scenarios: (1) over-reliance on single coarse-grained image regions, and (2) semantic fragmentation between successive reasoning steps. To address these issues, we propose the \textbf{CoCoT} (\textbf{Co}llaborative \textbf{Co}ross-modal \textbf{T}hought) framework, built upon two key innovations: a) Dynamic Multi-Region Grounding to adaptively detect the most relevant image regions based on the question, and b) Relation-Aware Reasoning to enable multi-region collaboration by iteratively aligning visual cues to form a coherent and logical chain of thought.
Through this approach, we construct the \textbf{CoCoT-70K} dataset, comprising 74,691 high-quality samples with multi-region annotations and structured reasoning chains. Extensive experiments demonstrate that CoCoT significantly enhances complex visual reasoning, achieving an average accuracy improvement of \textbf{15.4\%} on LLaVA-1.5 and \textbf{4.0\%} on Qwen2-VL across six challenging benchmarks. The data and code are available at: \url{https://github.com/deer-echo/CoCoT}.
 
\end{abstract}

\section{Introduction}
The Chain-of-Thought (CoT) paradigm has markedly advanced the reasoning capabilities of Large Language Models (LLMs) by generating sequential rationales~\cite{wei2022chain}. Multi-modal CoT, further serves as a vital bridge connecting visual perception with high-level reasoning~\cite{DBLP:conf/nips/ShaoQ0SZW0024}, and has become a cornerstone technique in Multimodal Large Language Models (MLLMs)~\cite{wang2024qwen2}.
By decomposing complex queries into structured steps grounded in visual evidence, 
multi-modal CoT methods have demonstrated strong performance across a spectrum of tasks including visual question answering, document analysis, and video reasoning~\cite{DBLP:journals/corr/abs-2504-18397, zhang2025chain, DBLP:journals/corr/abs-2503-05255, DBLP:conf/aaai/WangHHXLLS24, DBLP:conf/aaai/ChengCZ0FCL025}.

\begin{wrapfigure}[12]{r}{0.52\textwidth}
\centering
\includegraphics[width=\linewidth,trim=0 40 0 5, clip]{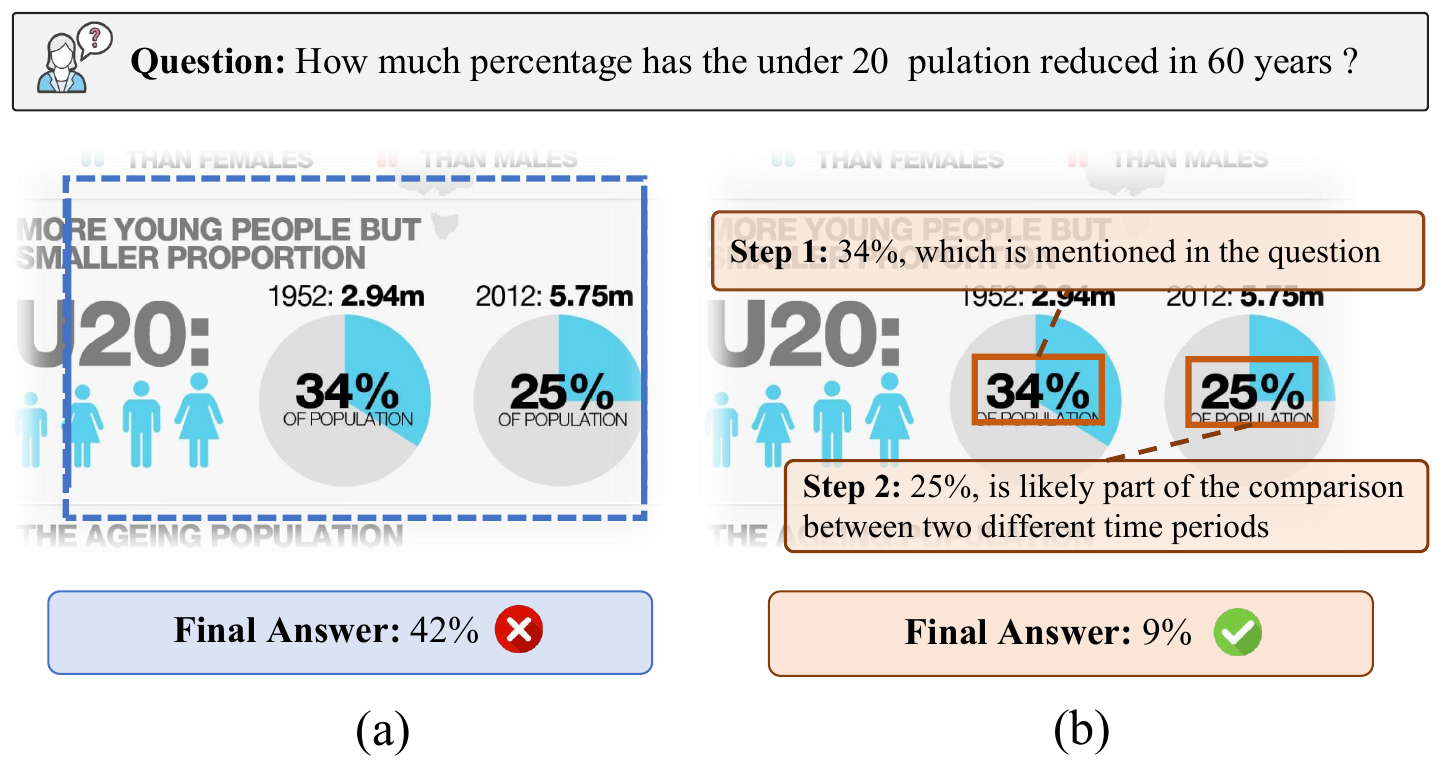}
\caption{Single-region CoT vs. CoCoT.}
\label{fig:example}
\end{wrapfigure}

Recent multi-modal CoT methods such as Visual CoT \cite{DBLP:conf/nips/ShaoQ0SZW0024} and SPHINX \cite{DBLP:conf/eccv/LinLZGQXQSCHHZHQL24} localize a critical region and generate a reasoning step based on isolated cue. While effective for simple queries, these approaches suffer from two fundamental limitations: (1) {over-reliance on single coarse-grained image regions}, and (2) {semantic fragmentation between successive reasoning steps}.
As illustrated in Fig.\ref{fig:example}a, recent models often generate a large bounding box, but this introduces excessive and irrelevant visual context, which dilutes critical information and harms performance.

\begin{figure}[tbp]
  \centering
  \includegraphics[width=\textwidth]{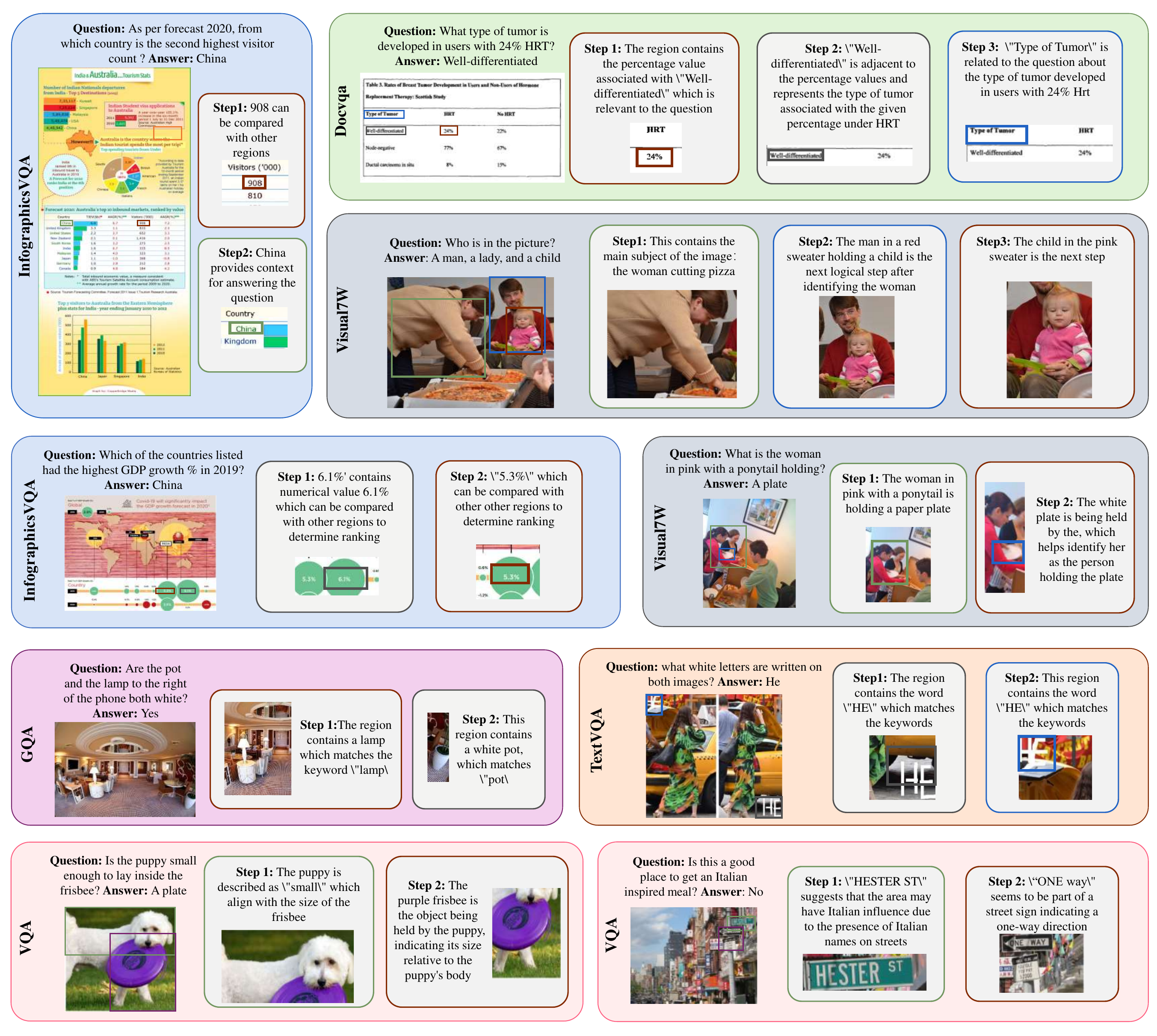}
  \caption{Examples of six datasets in the CoCoT-70K dataset.}
  \label{fig:dataset}
  \vspace{-5mm} 
\end{figure}

\begin{figure}[tbp]
  \centering
  \includegraphics[width=0.9\textwidth]{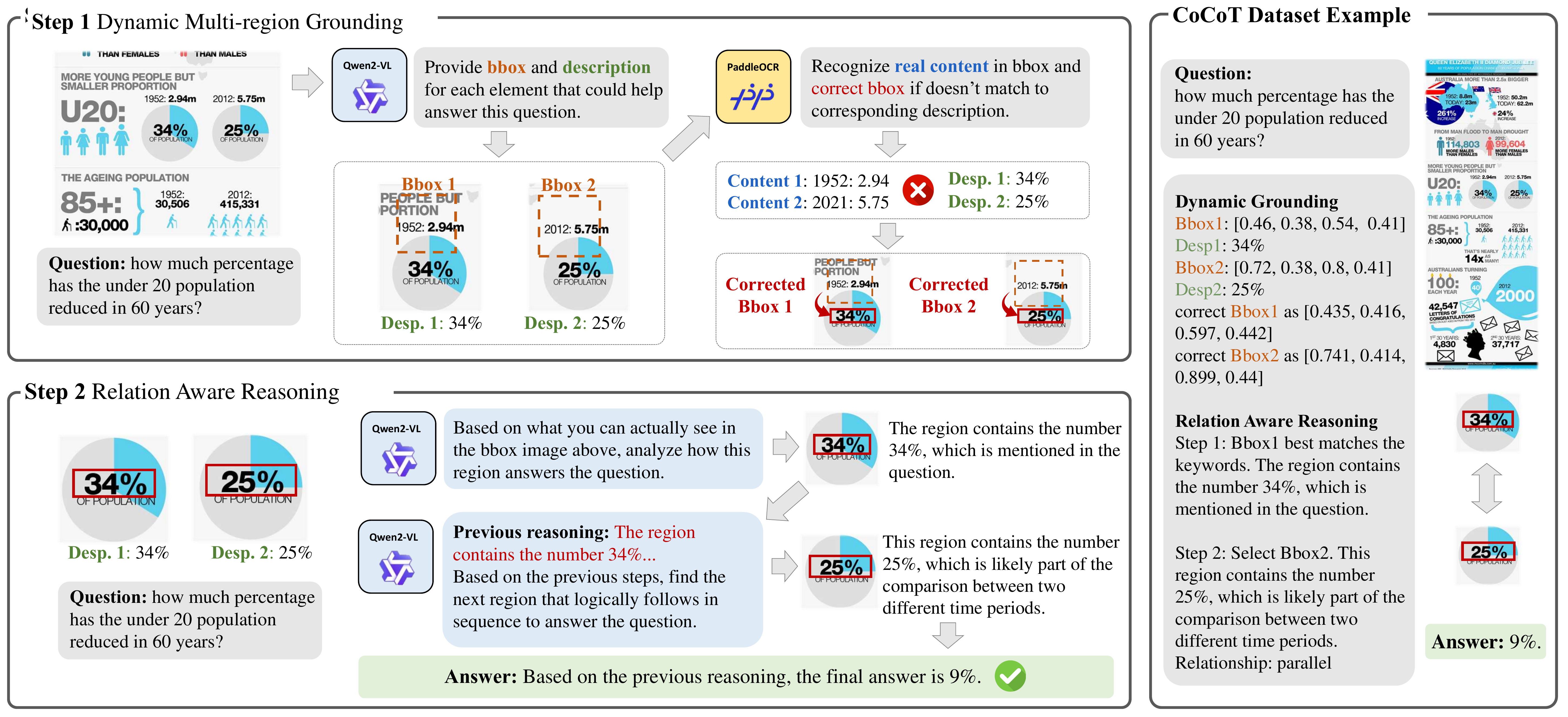}
  \caption{Overview of CoCoT. }
  \label{fig:overview}
\end{figure}

In contrast, human cognition operates through collaborative perception: we dynamically shift attention across multiple regions, bind them into semantic concepts, and infer relationships to form a holistic understanding. To bridge this gap, we propose the \textbf{CoCoT} (\textbf{Co}llaborative \textbf{Co}ross-modal \textbf{T}hought) framework, designed to directly address the two core limitations above. As shown in Fig.\ref{fig:overview}, CoCoT introduces: a) \textbf{Dynamic Multi-Region Grounding}: This component directly tackles the single-region reliance by collaborating with MLLMs and OCR to adaptively detect multiple precise regions most relevant to the question. b) \textbf{Relation-Aware Reasoning}: This process resolves semantic fragmentation by enabling multi-region collaboration to form a coherent and logical chain of thought, as shown in Fig.\ref{fig:example}b. To support this methodology, we construct the \textbf{CoCoT-70K} dataset (see examples in Fig.\ref{fig:dataset}), comprising 74,691 high-quality samples with multi-region annotations and structured reasoning chains. Extensive experiments demonstrate that CoCoT effectively overcomes the limitations of prior work, enabling significant improvements on complex visual reasoning tasks.

\vspace{2ex} 
\section{Overview of CoCoT}

\textbf{Dynamic Multi-Region Grounding.}
\label{sec:detection}
The recent one-region method, such as Visual CoT, generates a bounding box to indicate the model's attention to an image. This method fails to provide effective information for complex questions (e.g., Which of the countries listed had the highest GDP growth \% in 2019). 
In these tough conditions, the bounding box tends to be too small to get enough information or too big to distinguish effective information. 

Thus, we design a dynamic way that collaborates with Multimodal Large Language Model (MLLM) and Optical Character Recognition (OCR) to generate appropriate regions. 
Firstly, Qwen2-VL\cite{wang2024qwen2} is encouraged to generate multiple regions and the corresponding descriptions. The descriptions of this step always soundly match the question, while the bounding boxes are inaccurate. 
To correct these bounding boxes, secondly, we compare the content (extracted by PaddleOCR \cite{DBLP:journals/corr/abs-2009-09941}) with the description, if the content of the region can't match its descriptions, then we search for a better region, whose content is similar to description. If Qwen2-VL fail to give usable regions, the regions of keywords will be provided by OCR. This dynamic method combines the comprehension ability of Qwen2-VL and the localization precision of PaddleOCR, producing high-quality grounded representations that align textual semantics with visual spatial features. 

\begin{wrapfigure}[12]{t}{0.25\textwidth}
\centering
\includegraphics[width=\linewidth]{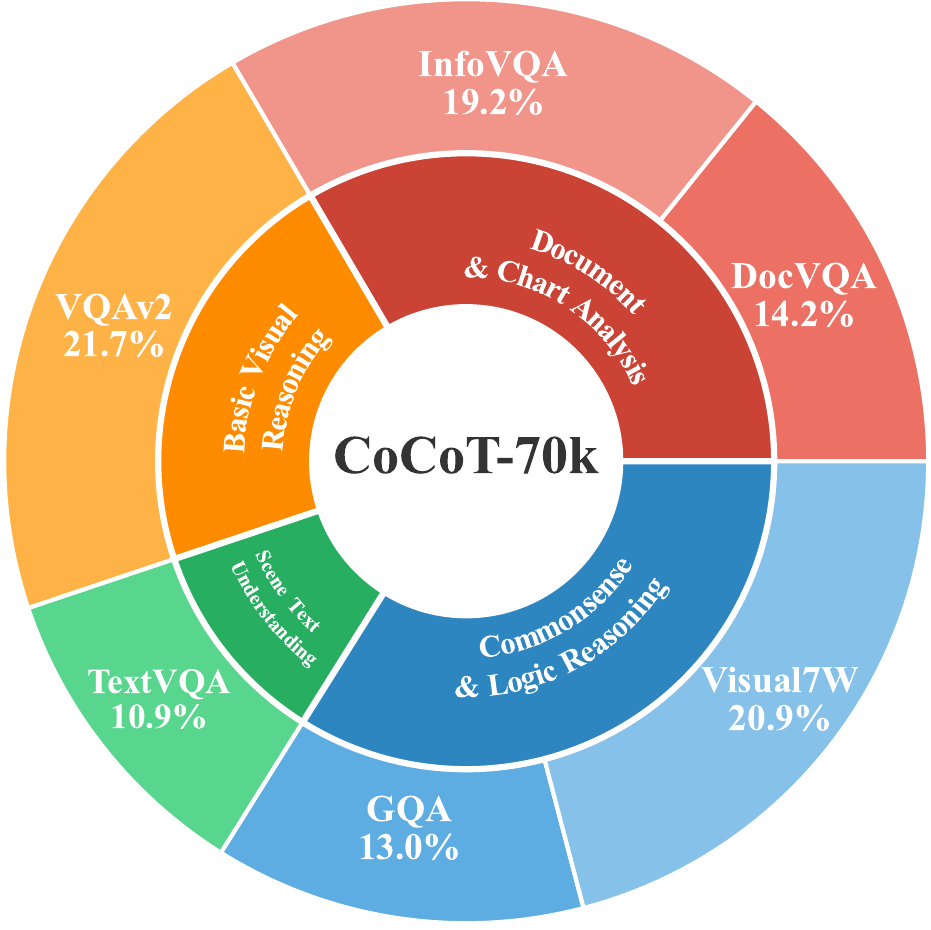}
\vspace{-4mm}
\caption{Data statistic of CoCoT-70k.}
\label{fig:statistic}
\end{wrapfigure}

\textbf{Relation-Aware Reasoning.}
\label{sec:reasoning}
We simulate human habits to construct this relation-aware reasoning process: First, \textbf{read the question}—parsing the problem into several keywords; Second, \textbf{locate keywords in the image} as entry points, where Qwen2-VL determine which bounding box generated in the grounding stage should be selected for positioning; Third, feed the selected region into Qwen2-VL  along with unselected regions, prompting Qwen2-VL to choose the next most relevant region and determine their relationship. Notably, region relationships are categorized into \textbf{parallel} and \textbf{sequential} types—parallel relationships generate logic chains like $A$→$B$, $C$→$D$, while sequential ones produce reasoning chains like $A$→$B$→$C$. The system iteratively inputs updated chains and candidate regions into Qwen2-VL, continuing until the current region sufficiently answers the question or all regions are exhausted.


\section{CoCoT-70k Dataset}

Based on the aforementioned pipeline, we construct CoCoT-70K, a high-quality dataset specifically designed for complex visual reasoning tasks. As shown in Fig.\ref{fig:statistic}, this dataset integrates six authoritative sources across four critical domains: \textbf{Basic Visual Reasoning}, \textbf{Document \& Chart Analysis}, \textbf{Commensense \& Logic Reasoning}, and \textbf{Scene Text Understanding}. This structured selection ensures broad coverage of essential visual-language capabilities: Basic visual reasoning tasks trains the fundamental ability to "see" and describe the explicit contents of a natural scene; document and chart data enhances structural understanding and precise information extraction from graphical and textual layouts; commensense and logic reasoning tasks develop deep visual commonsense and contextual inference in natural scenes; while text-rich image understanding fosters robust visual and semantic comprehension of embedded text. These six datasets are filtered to retain only samples with high keyword counts, thereby selecting for more complex and challenging questions (see Appendix for details). Then, we augment the original image-question-answer triplets with bounding boxes, region descriptions, and structured reasoning chains. The examples of the CoCoT-70k dataset are shown in Fig.\ref{fig:dataset}.

\section{Experiments}
\textbf{Experiment Setup.}
We evaluate our method on six multimodal QA benchmarks (InfographicsVQA \cite{mathew2022infographicvqa}, DocVQA\cite{mathew2021docvqa}, TextVQA \cite{singh2019towards}, Visual-7W \cite{zhu2016visual7w}, GQA \cite{hudson2019gqa}, and VQA-v2 \cite{goyal2017making}), comparing its performance against two baseline models (LLaVA-1.5 7B\cite{llava-1.5} and Qwen2-VL 7B \cite{wang2024qwen2}) and Visual CoT \cite{DBLP:conf/nips/ShaoQ0SZW0024} (a chain-of-thought-based visual reasoning model). To assess the capability in complex reasoning scenarios, we specifically select questions with dense keywords and multi-step or parallel-answer requirements (see Appendix for dataset details). Same as Visual CoT, our model is fine-tuned from LLaVA-1.5, epoch is 1, and batch is 256 for every fine-tuning stage. All experiments were conducted on a hardware setup with 4 NVIDIA V100 GPUs (32GB memory each), utilizing mixed-precision training for computational efficiency. 
Evaluation uses robust matching that extracts core answers from verbose responses and handles semantic equivalence, with separate analysis for single-bbox vs multi-bbox questions to assess complexity-dependent performance. 

\textbf{Main Results.}
As shown in Table \ref{tab:accuracy_comparison}, CoCoT-70k dataset is evaluated from two perspectives: inference and training. 
For inference, we compare three distinct methodologies:
Direct inference (generating answers directly from the question and original image),
CoCoT (first generating a reasoning chain using bounding boxes and descriptions, then producing the final answer), and
VisCoT-style inference (first giving a single bounding box, then generating the answer).
LLaVA-1.5 without specific training fails to effectively utilize the VisCoT method (i.e., it cannot produce valid bounding boxes). In contrast, our CoCoT chain-based reasoning consistently improves performance across all tasks, yielding an average accuracy gain of 15.4\%. For Qwen2-VL, our method generally delivers the strongest overall performance, although it is slightly outperformed by VisCoT on certain tasks (e.g., InfoVQA). We hypothesize that excessive detail in the reasoning chain may sometimes constrain stronger, generalist models like Qwen2-VL.

For training, we fine-tune the LLaVA-1.5 model using 20\% of our data (14k samples) in a two-stage procedure. Remarkably, this limited training set achieves performance comparable to VisCoT (which uses 363k fine-tuning samples based on LLaVA-1.5). However, it is important to note that our model underperforms in tasks requiring multi-box reasoning compared to the fully fine-tuned VisCoT, indicating that complex visual reasoning necessitates larger-scale training data.

\begin{table*}[t]
\caption{Accuracy comparison across datasets with single-box and multi-box samples. CoCoT means using our bounding boxes and corresponding descriptions to assist chain-based inference; VisCoT means applying two-satge inference as Visual CoT; * means training on annotated data before inference.}
\centering
\resizebox{1\linewidth}{!}{
\begin{tabular}{>{\centering\arraybackslash}l|>
{\centering\arraybackslash}c|>{\centering\arraybackslash}c>{\centering\arraybackslash}c|>{\centering\arraybackslash}c>{\centering\arraybackslash}c|>{\centering\arraybackslash}c>{\centering\arraybackslash}c|>{\centering\arraybackslash}c>{\centering\arraybackslash}c|>{\centering\arraybackslash}c>{\centering\arraybackslash}c|>{\centering\arraybackslash}c>{\centering\arraybackslash}c|>{\centering\arraybackslash}c>{\centering\arraybackslash}c>{\centering\arraybackslash}c}
\bottomrule

\toprule
\multirow{1}{*}{Method}&\parbox{1.2cm}{Training samples}& \multicolumn{2}{c|}{InfoVQA} & \multicolumn{2}{c|}{DocVQA} & \multicolumn{2}{c|}{TextVQA} & \multicolumn{2}{c|}{Visual7W} & \multicolumn{2}{c|}{GQA} & \multicolumn{2}{c|}{VQAv2} & \multicolumn{3}{c}{Average} \\
\cline{2-16}
& & Single & Multi & Single & Multi & Single & Multi & Single & Multi & Single & Multi & Single & Multi & Single & Multi & Overall \\

\toprule
LLaVA-1.5 & 0 & 16.3 & 19.5 & 22.3 & 13.6 & 13.5 & 14.6 & 24.8 & 23.7 & 62.6 & 52.6 & 29.2 & 28.8 & 25.7 & 28.9 & 27.0  \\
CoCoT (LLaVA) &0& 48.8 & 38.5 & 43.0 & 50.0 & 47.6 & 43.7 & 31.5 & 32.7 & 51.7 & 38.1 & 49.8 & 37.8 &45.4 &38.4  & 42.4 \\
VisCoT* (LLaVA) &363k & 28.8 & 22.4 & 41.0 & 48.9 & 51.9 & 39.7 & 43.2 & 41.0 & 79.1 & 85.8 & 51.9 & 46.8 & 47.6 & 49.8 & 48.5  \\
CoCoT* (LLaVA) &14k & 49.8 & 28.8 & 45.6 & 50.0 & 53.3 & 35.1 & 
45.0 & 32.0 & 60.7 & 61.9 & 52.4 & 43.1 & 50.6 & 42.2 & 47.0 \\
\midrule
Qwen2-VL &0& 76.9 & 81.5 & 96.4 & 93.2 & 71.3 & 70.2 & 41.9 & 36.7 & 77.7 & 82.4 & 63.1 & 53.9 & 74.2 & 65.6 & 70.5 \\
CoCoT (Qwen) &0& 81.0 & 78.0 & 95.6 & 92.0 & 75.9 & 71.5 & 53.2 & 54.3 & 82.5 & 82.0 & 64.8 & 58.4  & 77.9 & 69.9 & 74.5\\
VisCoT (Qwen) &0& 80.7 & 84.4 & 97.6 & 93.2 & 71.9 & 68.2 & 41.9 & 36.0 & 84.8 & 88.9 & 58.8 & 54.7 & 75.5 & 64.7 & 72.0 \\
\bottomrule

\toprule
\end{tabular}
}
\label{tab:accuracy_comparison}
\vspace{-10mm}
\end{table*}

\section{Conclusion}
In this work, we propose CoCoT to address semantic fragmentation in multi-modal reasoning. Our framework introduces dynamic multi-region grounding and relation-aware reasoning, along with the CoCoT-70K dataset. Experiments demonstrate consistent improvements across benchmarks.

\bibliographystyle{plainnat}  
\bibliography{reference}

\appendix







\title{Appendix}

\section{Framework details}

\subsection{Model details}
Our base model setup aligns with Visual CoT \cite{DBLP:conf/nips/ShaoQ0SZW0024}. Specifically, we employ the pre-trained CLIP ViT-L\/14\-336 as the vision encoder and Vicuna\-7B\-v1.5 as the large language model (LLM), which exhibits stronger instruction-following capabilities in linguistic tasks compared to LLaMA. For an input image, we first use the vision encoder to extract visual features. Following the practice of LLaVA, we then project these image features into the word embedding space via a simple linear layer (mlp2x\_gelu), obtaining visual tokens that match the dimensionality of the LLM. 
Based on these settings, we employ a novel two-stage progressive training paradigm where the first stage focuses on reasoning chain generation from multi-modal inputs (original image, bbox-cropped regions and question descriptions), while the second stage synthesizes final answers based on the generated reasoning chains and original images.

\subsection{Training Strategy and Implementation Details}
\label{sec:training}

To enhance the model's long-chain reasoning capabilities, we propose a two-stage progressive training framework for visual reasoning. Different from recent two-stage methods like mm-CoT \cite{DBLP:journals/tmlr/0001Z00KS24} which only use the original image, we decompose each training instance into separate samples for questions with multiple relevant bounding boxes. Each sample focuses on one specific region while maintaining global context through the original image, enabling the model to learn fine-grained region-specific reasoning patterns from multiple visual perspectives for the same question.

\textbf{Stage 1: Reasoning Chain Generation.} The model learns to generate detailed reasoning chains by processing original images paired with individual cropped regions, questions, and region descriptions. We train for 1 epoch with a learning rate of 2e-5 and batch size of 64 (achieved via 1 sample per device × 64 gradient accumulation steps). For image preprocessing, we adopt the crop-and-pad strategy from Visual CoT \cite{DBLP:conf/nips/ShaoQ0SZW0024}, which maintains aspect ratios while ensuring uniform 336×336 input dimensions, preserving spatial relationships crucial for accurate bounding box coordinate generation.

\textbf{Stage 2: Answer Synthesis.} We first use the trained Stage 1 model to generate reasoning chains for all training data through parallel inference across 4 GPUs. The model is then fine-tuned for 1 epoch on final answer synthesis using original images, questions, and the generated reasoning chains, employing a lower learning rate of 1e-5 with the same batch size configuration.

We randomly sample 20\% of the total CoCoT dataset (14,392 samples from six datasets) for pretraining. The Adam optimizer with zero weight decay and a cosine learning rate scheduler are utilized throughout. To conserve GPU memory during fine-tuning, we employ DeepSpeed ZeRO-3 with FP16 precision training. All models are trained using 4 × Tesla V100-32GB GPUs.

\subsection{Dataset details}

We curated six benchmark datasets by applying two filtering criteria: (1) questions containing multiple keywords (thresholds varying by dataset from >3 to >6 keywords) and (2) answers requiring compositional reasoning (containing conjunctions or multiple elements), the details are shown in Tab.\ref{tab:dataset_composition}. This process yielded 74,691 complex question-answer pairs that better simulate real-world visual reasoning challenges. For each dataset, 500 samples are randomly extracted to constitute the test set, with 20\% of the remaining samples then being allocated to form the model's training set.

\begin{table}[h]
\label{dataset}
\centering
\caption{CoCoT Dataset Composition}
\label{tab:dataset_composition}
\begin{tabularx}{\linewidth}{>{\RaggedRight}p{1.5cm}>{\centering}m{1.2cm}>{\RaggedRight}m{2cm}>{\RaggedRight}m{1.5cm}X}
\toprule
\textbf{Dataset} & \textbf{Samples} & \textbf{Filter Criteria} & \textbf{Multi Region Ratio} &  \textbf{Source Files} \\
\midrule
GQA \cite{hudson2019gqa} & 9,740 & 
Keywords >6 &
41.4\% &
\texttt{GQA\_val\_balanced.json} 
\texttt{GQA\_val\_all.json}
\texttt{GQA\_train\_balanced.json} \\
\hline
DocVQA \cite{mathew2021docvqa} & 10,650 & 
Keywords >4 or answers with ",/and" &
18.1\% &
\texttt{docvqa\_train\_reordered.jsonl}
\texttt{docvqa\_train\_v1.0\_reordered.json} \\
\hline
InfoVQA \cite{mathew2022infographicvqa} & 14,421 & 
Keywords >4 or parallel answers &
39.1\% &
\texttt{infographicVQA\_train\_v1.0.json}
\texttt{infographicVQA\_val\_v1.0.json} \\
\hline
TextVQA \cite{singh2019towards} & 8,205 & 
Keywords >3 or conjunction answers &
31.2\% &
\texttt{TextVQA\_train.json} \\
\hline
Visual7W \cite{zhu2016visual7w} & 15,675 & 
Keywords >3 or multi-part answers &
51.5\% &
\texttt{Visual7W\_telling.json} \\
\hline
VQAv2 \cite{goyal2017making} & 16,270 & 
Keywords >5 or compound answers &
54.5\% &
\texttt{VQA\_v2\_train.json} \\
\bottomrule

\end{tabularx}
\end{table}

\section{Ablation Study}

To investigate the effectiveness of Relation-Aware Reasoning, we conduct several ablation studies:
\begin{itemize}
    \item -RAR: Since the Relation-Aware Reasoning stage takes the description and bounding boxes as input to generate a reasoning chain, we simulate the absence of this module during inference by using only the description and bounding boxes.
    \item Replaced RAR: To explore whether relations can be directly generated in the Dynamic Multi-Region Grounding stage, we generate both multiple bounding boxes and their corresponding relationst o the question for each box, thereby replacing the reasoning chain.
    \item Qwen RAR: To examine the potential of the Relation-aware Reasoning stage, we directly use the chain generated by Qwen2-VL for inference.
    \item CoCoT: denotes the method where a reasoning chain is first generated based on the provided bounding boxes and description, and then the final answer is derived using this chain.
\end{itemize}
These four methods are compared against the direct inference baselines of LLaVA-1.5 and Qwen2-VL in Table \ref{tab:ablation}, with their performance changes reported relative to the baselines.

Experimental results demonstrate that Qwen2-VL-generated reasoning chains yield the most substantial performance improvement for LLaVA-1.5, achieving a significant accuracy gain of 23\%. For Qwen2-VL itself, our proposed method delivers optimal results with a 4\% accuracy improvement. In contrast, the Replaced RAR approach exhibits the worst performance across both baseline models, particularly reducing accuracy by 22.4\% on Qwen2-VL. This evidence indicates that jointly generating bounding boxes alongside relational rankings during the Dynamic Multi-Region Grounding stage is substantially less effective than decoupling these operations through a dedicated Relation-Aware Reasoning module, thereby validating the necessity of our proposed two-stage reasoning paradigm.

Furthermore, while the -RAR strategy shows competitive performance for LLaVA-1.5, it severely degrades Qwen2-VL's accuracy by 19.5\%. This contrasting behavior suggests that providing only region descriptions without explicit relational reasoning can enhance weaker models but critically impairs the capability of more advanced vision-language models, highlighting the importance of architectural compatibility with model capacity.

\begin{table*}[t]
\caption{Accuracy comparison across datasets with single-box and multi-box samples. Green (+) indicates improvement over Direct method, red (-) indicates decrease.}
\centering
\resizebox{1\linewidth}{!}{
\begin{tabular}{>{\centering\arraybackslash}l|>{\centering\arraybackslash}c>{\centering\arraybackslash}c|>{\centering\arraybackslash}c>{\centering\arraybackslash}c|>{\centering\arraybackslash}c>{\centering\arraybackslash}c|>{\centering\arraybackslash}c>{\centering\arraybackslash}c|>{\centering\arraybackslash}c>{\centering\arraybackslash}c|>{\centering\arraybackslash}c>{\centering\arraybackslash}c|>{\centering\arraybackslash}c>{\centering\arraybackslash}c>{\centering\arraybackslash}c}
\toprule
Method & \multicolumn{2}{c|}{infographics}& \multicolumn{2}{c|}{docvqa} & \multicolumn{2}{c|}{textvqa} & \multicolumn{2}{c|}{visual-7w} & \multicolumn{2}{c|}{GQA} & \multicolumn{2}{c|}{VQA-v2} & \multicolumn{3}{c}{Average} \\
\cmidrule(lr){2-3} \cmidrule(lr){4-5} \cmidrule(lr){6-7} \cmidrule(lr){8-9} \cmidrule(lr){10-11} \cmidrule(lr){12-13} \cmidrule(lr){14-16}
 & Single & Multi & Single & Multi & Single & Multi & Single & Multi & Single & Multi & Single & Multi & Single & Multi & Overall \\
\midrule
LLaVA-1.5  & 16.3 & 19.5 & 22.3 & 13.6 & 13.5 & 14.6 & 24.8 & 23.7 & 62.6 & 52.6 & 29.2 & 28.8 & 25.7 & 28.9 & 27.0 \\
LLaVA-1.5 (-RAR) & \textcolor{green}{+}28.4 & \textcolor{green}{+}7.3 & \textcolor{green}{+}20.9 & \textcolor{green}{+}18.2 & \textcolor{green}{+}39.8 & \textcolor{green}{+}21.2 & \textcolor{green}{+}22.5 & \textcolor{green}{+}11.2 & \textcolor{red}{-}0.5 & \textcolor{green}{+}6.9 & \textcolor{green}{+}22.7 & \textcolor{green}{+}8.7 & \textcolor{green}{+}23.8 & \textcolor{green}{+}10.7 & \textcolor{green}{+}18.3 \\
LLaVA-1.5 (Replaced RAR) & \textcolor{green}{+}12.9 & \textcolor{green}{+}3.4 & \textcolor{green}{+}10.0 & \textcolor{green}{+}9.1 & \textcolor{green}{+}13.1 & \textcolor{green}{+}3.3 & \textcolor{green}{+}13.5 & \textcolor{green}{+}6.9 & \textcolor{red}{-}0.5 & \textcolor{red}{-}2.1 & \textcolor{green}{+}8.6 & \textcolor{green}{+}4.2 & \textcolor{green}{+}10.1 & \textcolor{green}{+}3.4 & \textcolor{green}{+}7.3 \\
LLaVA-1.5 (Qwen RAR) & \textcolor{green}{+}29.8 & \textcolor{green}{+}13.2 & \textcolor{green}{+}33.3 & \textcolor{green}{+}36.4 & \textcolor{green}{+}44.7 & \textcolor{green}{+}34.4 & \textcolor{green}{+}20.2 & \textcolor{green}{+}16.2 & \textcolor{green}{+}5.6 & \textcolor{green}{+}2.4 & \textcolor{green}{+}21.9 & \textcolor{green}{+}13.5 & \textcolor{green}{+}28.4 & \textcolor{green}{+}15.5 & \textcolor{green}{+}23.0 \\
LLaVA-1.5 (CoCoT) & \textcolor{green}{+}32.5 & \textcolor{green}{+}19.0 & \textcolor{green}{+}20.7 & \textcolor{green}{+}36.4 & \textcolor{green}{+}34.1 & \textcolor{green}{+}29.1 & \textcolor{green}{+}6.7 & \textcolor{green}{+}9.0 & \textcolor{red}{-}10.9 & \textcolor{red}{-}14.5 & \textcolor{green}{+}20.6 & \textcolor{green}{+}9.0 & \textcolor{green}{+}19.7 & \textcolor{green}{+}9.5 & \textcolor{green}{+}15.4 \\
\midrule
Qwen2-VL & 76.9 & 81.5 & 96.4 & 93.2 & 71.3 & 70.2 & 41.9 & 36.7 & 77.7 & 82.4 & 63.1 & 53.9 & 74.2 & 65.6 & 70.5 \\
Qwen2-VL (-RAR) & \textcolor{red}{-}22.3 & \textcolor{red}{-}47.8 & \textcolor{red}{-}38.4 & \textcolor{red}{-}59.1 & \textcolor{red}{-}10.0 & \textcolor{red}{-}26.5 & \textcolor{red}{-}5.0 & \textcolor{red}{-}8.3 & \textcolor{red}{-}8.0 & \textcolor{red}{-}8.4 & \textcolor{red}{-}10.7 & \textcolor{red}{-}13.5 & \textcolor{red}{-}18.2 & \textcolor{red}{-}21.3 & \textcolor{red}{-}19.5 \\
Qwen2-VL (Replaced RAR) & \textcolor{red}{-}33.5 & \textcolor{red}{-}48.8 & \textcolor{red}{-}44.5 & \textcolor{red}{-}64.8 & \textcolor{red}{-}20.3 & \textcolor{red}{-}31.8 & \textcolor{red}{-}8.6 & \textcolor{red}{-}6.5 & \textcolor{red}{-}7.1 & \textcolor{red}{-}4.9 & \textcolor{red}{-}10.7 & \textcolor{red}{-}8.6 & \textcolor{red}{-}24.0 & \textcolor{red}{-}20.3 & \textcolor{red}{-}22.4 \\
Qwen2-VL (Qwen RAR) & \textcolor{green}{+}0.7 & \textcolor{red}{-}4.4 & \textcolor{green}{+}0.7 & \textcolor{green}{+}1.1 & \textcolor{green}{+}0.6 & \textcolor{red}{-}2.7 & \textcolor{green}{+}5.8 & \textcolor{green}{+}6.1 & \textcolor{green}{+}6.7 & \textcolor{green}{+}0.3 & \textcolor{red}{-}0.9 & \textcolor{red}{-}0.3 & \textcolor{green}{+}1.8 & \textcolor{green}{+}0.4 & \textcolor{green}{+}1.3 \\
Qwen2-VL (CoCoT) & \textcolor{green}{+}4.1 & \textcolor{red}{-}3.5 & \textcolor{red}{-}0.8 & \textcolor{red}{-}1.2 & \textcolor{green}{+}4.6 & \textcolor{green}{+}1.3 & \textcolor{green}{+}11.3 & \textcolor{green}{+}17.6 & \textcolor{green}{+}4.8 & \textcolor{red}{-}0.4 & \textcolor{green}{+}1.7 & \textcolor{green}{+}4.5 & \textcolor{green}{+}3.7 & \textcolor{green}{+}4.3 & \textcolor{green}{+}4.0 \\
\bottomrule
\end{tabular}
}
\label{tab:ablation}
\vspace{-3mm}
\end{table*}

\section{Prompt design}
Our approach employs a multi-stage prompting strategy to construct comprehensive reasoning chains for visual question answering. The key innovation lies in our question-type-aware reasoning chain construction, which automatically distinguishes between sequential reasoning (A→B→C) and parallel evidence gathering (A→B; A→C) based on question analysis.

\textbf{Generation Stage:} We design adaptive prompts that handle single-bbox and multi-bbox scenarios differently. For multi-bbox cases, our iterative prompts incorporate spatial relationship analysis, guiding the model to explore regions in similar positions (same row/column) for parallel questions, while ensuring comprehensive region exploration through progress tracking.

\textbf{Training and Inference:} We implement a two-stage progressive framework where Stage 1 generates reasoning chains from visual regions and descriptions, and Stage 2 synthesizes final answers. During inference, we evaluate six distinct strategies including our method, ablation studies, and comparisons with Visual CoT, enabling comprehensive analysis of different reasoning approaches.

Table~\ref{tab:all_prompts} presents the complete prompt templates, demonstrating our systematic design for effective multi-modal reasoning chain construction.

\begin{table}[h]
\centering
\caption{Bbox Generation and Reasoning Chain Construction Prompts}
\label{tab:all_prompts}
\begin{tabular}{|p{2.5cm}|p{10cm}|}
\hline
\textbf{Task} & \textbf{Prompt Template} \\
\hline
\hline
\multicolumn{2}{|c|}{\textbf{GENERATION STAGE PROMPTS}} \\
\hline
\textbf{Single-Step} & \texttt{Question: \{question\}} \\
\textbf{Reasoning} & \texttt{Keywords: \{keywords\}} \\
\textbf{Chain} & \texttt{Available region: Region 0: \{bbox\_content\}} \\
& \texttt{Task: Analyze how this region answers the question. Generate a concise explanation (max 30 words).} \\
& \texttt{IMPORTANT: Base your analysis ONLY on what you can actually see in the bbox image above.} \\
& \texttt{Output format: SELECTED\_REGION: Region 0, ROLE: direct\_answer/evidence, REASONING: [key info] directly answers/provides [question aspect], RELATIONSHIP: none} \\
\hline
\textbf{Multi-Step} & \texttt{Question: \{question\}} \\
\textbf{Reasoning} & \texttt{Progress: Used \{used\_count\}/\{total\_count\} regions. Try to explore most regions before concluding.} \\
\textbf{Chain} & \texttt{Question Type Analysis: \{question\_type\} (Sequential: A->B->C; Parallel: A->B; A->C)} \\
& \texttt{Previous reasoning steps: [previous steps]} \\
& \texttt{Available regions for this step: [available regions]} \\
& \texttt{Task: \{role\_instruction\}} \\
& \texttt{Output format: SELECTED\_REGION: [Region X], ROLE: [keyword\_match/evidence/conclusion], REASONING: [explanation], RELATIONSHIP: [sequential/parallel/none]} \\
\hline
\textbf{Training} & \textbf{Stage 1:} \texttt{Question: \{question\}, Description: \{description\}} \\
\textbf{Stages} & \texttt{Based on the image and highlighted region, provide a step-by-step reasoning chain to answer the question:} \\
& \textbf{Stage 2:} \texttt{Question: \{question\}, Reasoning Chain: \{chain\_text\}} \\
& \texttt{Based on the reasoning chain, provide the final answer:} \\
\hline
\hline
\multicolumn{2}{|c|}{\textbf{INFERENCE STAGE PROMPTS}} \\
\hline
\textbf{Direct} & \texttt{\{question\}} \\
\textbf{Inference} & \textit{(No additional prompt, uses original question directly)} \\
\hline
\textbf{Two-Stage} & \textbf{My Method Stage 1:} \texttt{Based on the description '\{description\}', analyze this image region and provide relevant information for answering: \{question\}} \\
\textbf{Methods} & \textbf{My Method Stage 2:} \texttt{Question: \{question\}, Based on the following analysis: \{chain\_context\}, Provide the final answer:} \\
& \textbf{Visual CoT Stage 1:} \texttt{<image> \{question\} Please provide the bounding box coordinate of the region this question asks about.} \\
& \textbf{Visual CoT Stage 2:} \texttt{<image>} \textit{(Uses cropped bbox region to answer original question)} \\
\hline
\textbf{Single-Stage} & \textbf{-RAR:} \texttt{Description: \{description\}, Question: \{question\}, Answer:} \\
\textbf{Methods} & \textbf{Replaced RAR:} \texttt{Content: \{content\_relation\}, Question: \{question\}, Answer:} \\
& \textbf{Qwen RAR:} \texttt{Chain: \{chain\_text\}, Question: \{question\}, Based on the chain, provide the answer:} \\
\hline
\end{tabular}
\end{table}

\section{Limitations}

The CoCoT-70k dataset presented in this paper is constructed through a two-stage pipeline. In the first stage, multiple relevant regions are identified using Qwen2-VL and PaddleOCR. In the second stage, Qwen2-VL is employed to sort these regions, determine their interrelations, and generate the corresponding reasoning chains. Although PaddleOCR is used for post-processing correction, the regions extracted in the first stage remain imperfect—particularly in datasets such as GQA, where very few textual elements can be successfully recognized by PaddleOCR. This observation indicates that generating high-quality multimodal reasoning chains strongly depends on robust visual perception capabilities.

Furthermore, during training, our approach only fine-tunes the model to generate reasoning chains and subsequently produce final answers based on those chains. The stage of generating bounding boxes is not included in the training process, as LLaVA-based models struggle to produce multiple regions accurately in a single pass. We anticipate addressing this limitation in future work by employing vision models with stronger localization capabilities.




\end{document}